\documentclass[11pt,a4paper]{article}
\usepackage[hyperref]{acl2020}
\usepackage{times}
\usepackage{etex}  
\usepackage{standalone}
\usepackage{latexsym}
\usepackage{url}
\usepackage{amsmath}
\usepackage{amssymb}
\usepackage{amsfonts}
\usepackage{graphicx}
\usepackage[skip=5pt]{caption}
\usepackage{subcaption}
\usepackage{array}
\usepackage{tabu}
\usepackage{makecell}
\usepackage{subcaption}
\usepackage{paralist}
\usepackage{cases}
\usepackage{diagbox}
\usepackage{enumitem}
\usepackage{soul}
\usepackage{multirow}
\usepackage{verbatim}
\usepackage{tabulary}
\usepackage{booktabs}
\usepackage[mathscr]{euscript}
\usepackage{mathtools}
\usepackage{amssymb}
\usepackage{algorithm}
\usepackage{algpseudocode}
\usepackage{stmaryrd}
\usepackage{amsthm}
\usepackage{tikz-dependency}
\usetikzlibrary{automata,decorations.markings,arrows,positioning,matrix,calc,patterns,angles,quotes,calc}
\usepackage{adjustbox}
\usepackage[toc,page]{appendix}
\usepackage{tabularx}
\usepackage{xspace}
\usepackage[toc,page]{appendix}
\usepackage{tabulary}
\usepackage{afterpage}

\aclfinalcopy
\algrenewcommand{\algorithmiccomment}[1]{\leavevmode$\triangleright$ #1}

\usepackage{color}
\usepackage{bm}
\definecolor{orange}{rgb}{1,0.5,0}
\definecolor{mdgreen}{rgb}{0.05,0.6,0.05}
\definecolor{mdblue}{rgb}{0,0,0.7}
\definecolor{dkblue}{rgb}{0,0,0.5}
\definecolor{dkgray}{rgb}{0.3,0.3,0.3}
\definecolor{slate}{rgb}{0.25,0.25,0.4}
\definecolor{gray}{rgb}{0.5,0.5,0.5}
\definecolor{ltgray}{rgb}{0.7,0.7,0.7}
\definecolor{purple}{rgb}{0.7,0,1.0}
\definecolor{lavender}{rgb}{0.65,0.55,1.0}

\definecolor{mypurple}{RGB}{111,61,121}
\definecolor{myblue}{RGB}{46,88,180}
\definecolor{myred}{RGB}{181,68,106}
\definecolor{myyellow}{RGB}{204,143,55}

\newcommand{\ensuretext}[1]{#1}

\newcommand{\arkcomment}[3]{\ensuretext{\textcolor{#3}{[#1 #2]}}}
\renewcommand{\arkcomment}[3]{}  

\newcommand{\figref}[1]{Figure~\ref{fig:#1}}

\newcommand{\term}[1]{\textbf{#1}} 

\newcommand{\interalia}[1]{\citep[\emph{inter alia}]{#1}}

\newcommand{\multihead}{\operatorname{MultiHead}}

\newcommand{\softmax}{\operatorname{softmax}}

\newcolumntype{L}[1]{>{\raggedright\let\newline\\\arraybackslash\hspace{0pt}}m{#1}}
\newcolumntype{C}[1]{>{\centering\let\newline\\\arraybackslash\hspace{0pt}}m{#1}}
\newcolumntype{R}[1]{>{\raggedleft\let\newline\\\arraybackslash\hspace{0pt}}m{#1}}

\theoremstyle{definition}

\theoremstyle{remark}

\setul{1pt}{.4pt}

\DeclareCaptionFont{tiny}{\tiny}

\usepackage{url}
\usepackage[shortcuts]{extdash}  

\usepackage{blindtext}
\usepackage{graphicx}
\usepackage{capt-of}
\usepackage{booktabs}
\usepackage{varwidth}
\newsavebox\tmpbox

\def\vf{{\mathbf{f}}}
\def\vg{{\mathbf{g}}}

\def\vx{{\mathbf{x}}}

\def\mH{{\mathbf{H}}}

\def\mK{{\mathbf{K}}}

\def\mQ{{\mathbf{Q}}}

\def\mV{{\mathbf{V}}}
\def\mW{{\mathbf{W}}}
\def\mX{{\mathbf{X}}}

\def\mZ{{\mathbf{Z}}}

\def\gD{{\mathcal{D}}}

\def\gL{{\mathcal{L}}}

\newcommand{\ourmodel}{MAE\xspace}

\newcommand{\ourmodelsc}{\textsc{Mae-7}\xspace}
\newcommand{\ourmodeltsc}{\textsc{Mae-6}\xspace}
\newcommand{\joint}{\textsc{noBCD}\xspace}

\newcommand{\base}{\textsc{Base}\xspace}
\newcommand{\moe}{\operatorname{MoE}}
\newcommand{\headdrop}{\textsc{uni-Mae-7}\xspace}
\newcommand{\headdropt}{\textsc{uni-Mae-6}\xspace}
\newcommand{\scratch}{\textsc{Scratch}\xspace}
\newcommand{\finetuneg}{\textsc{FtG}\xspace}
\newcommand{\finetunegp}{\textsc{FtG+}\xspace}
\newcommand{\finetuneall}{\textsc{FtAll}\xspace}
\newcommand{\nofinetune}{\textsc{NoFt}\xspace}

\title{A Mixture of $h-1$ Heads is Better than $h$ Heads}

\author{Hao Peng$^\spadesuit$ \quad Roy Schwartz$^{\diamondsuit\spadesuit}$ \quad Dianqi Li$^\clubsuit$  \quad Noah A. Smith$^{\diamondsuit\spadesuit}$ \\
  $^\diamondsuit$Allen Institute for Artificial Intelligence\\ 
  $^\spadesuit$Paul G. Allen School of Computer Science \& Engineering,
  University of Washington\\ 
  $^\clubsuit$Department of Electrical \& Computer Engineering,
  University of Washington\\ 
  {\tt \{hapeng,roysch,nasmith\}@cs.washington.edu, dianqili@uw.edu}
}

\date{}

\begin{document}
	\setlength{\abovedisplayskip}{5pt}
	\setlength{\belowdisplayskip}{5pt}
	\maketitle
	\begin{abstract}
Multi-head attentive neural architectures have achieved state-of-the-art results 
on a variety of natural language processing tasks. 
Evidence has shown that they are overparameterized; attention heads
can be pruned without significant performance loss. 
In this work, we instead ``reallocate'' them---the model 
learns to activate different heads on different inputs.
Drawing connections between multi-head attention
and mixture of experts, we
propose the \term{m}ixture of \term{a}ttentive \term{e}xperts model (\ourmodel).
\ourmodel is trained using
a block coordinate descent algorithm 
that alternates between updating
(1) the responsibilities of the experts and 
(2) their parameters. 
Experiments on machine translation and language modeling show that
\ourmodel~outperforms strong baselines on both tasks.
Particularly, on the WMT14 English to German translation dataset, 
\ourmodel~improves over ``transformer-base'' by 0.8 BLEU, with a comparable number of parameters. 
Our analysis shows 
that our model learns to specialize different experts to different inputs.\footnote{Our implementation is publicly available at \url{https://github.com/Noahs-ARK/MAE}.}
\end{abstract}

	\section{Introduction}

\label{sec:introduction}
The transformer architecture and its variants
achieve state-of-the-art performance
across a variety of NLP tasks,
including machine translation~\citep{vaswani2017attention,ott2018scaling},
language modeling~\citep{radford2018language,baevski2018adaptive},
semantic role labeling~\citep{strubell2018linguistically},
and more~\citep{delvin2019bert,liu2019roberta,yang2019xlnet}.
Under the hood, multi-head attention provides the driving force:
multiple separately parameterized attention functions 
act in parallel to contextualize the input representations;
their outputs are then gathered by an affine transformation,
and fed to onward computation.

\begin{figure}
\centering
\includegraphics[trim={5cm 7.3cm 5cm 7.3cm},clip,width=\columnwidth]{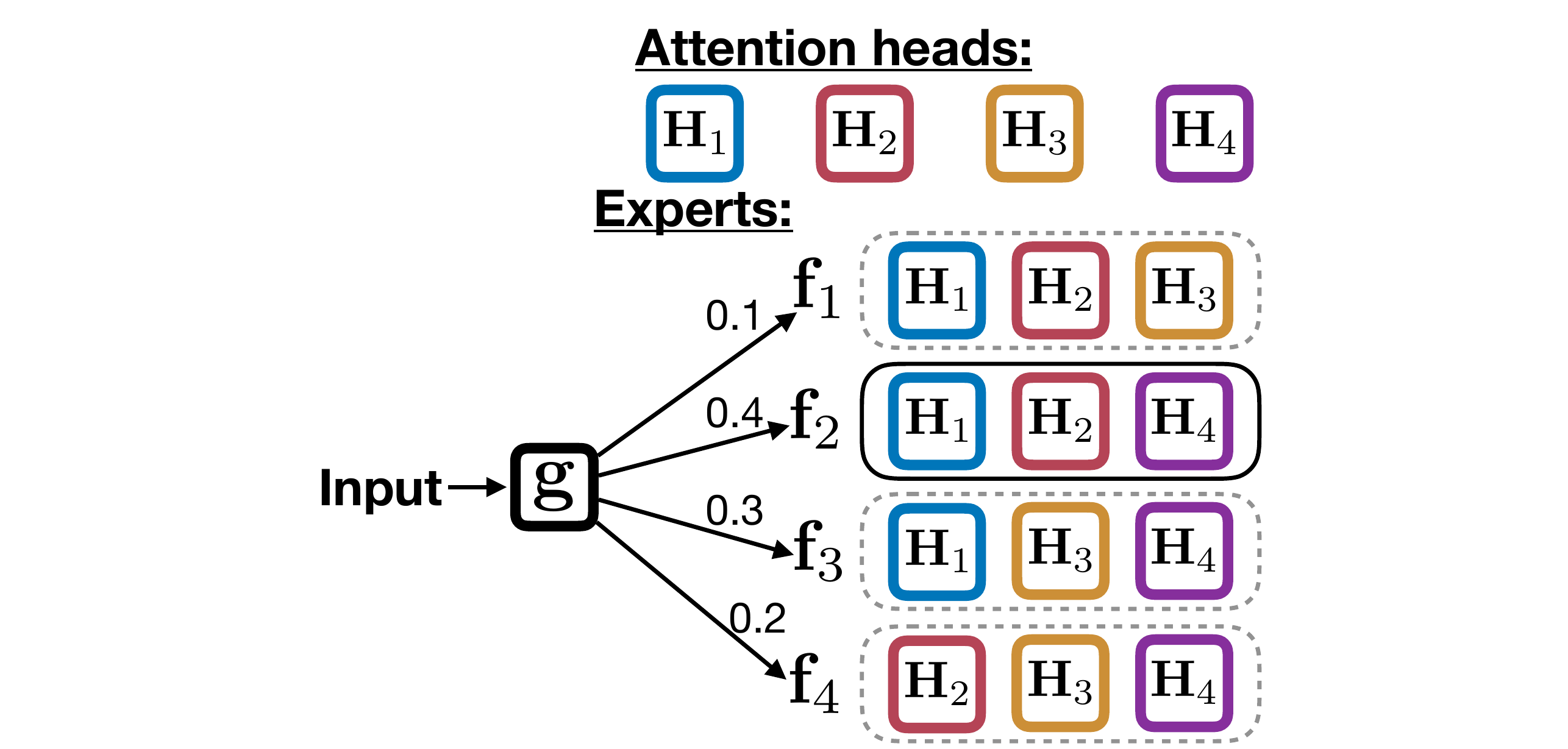}
\caption{\label{fig:model} Illustration of \ourmodel: a mixture of attentive experts.
Each $\mH_i$ box is an attention head in a given layer; there are $h$ of them in total. 
Experts are groups of $h-1$ attention heads.
\ourmodel learns an input-dependent distribution of the experts ($\vg$).
At each training step, a single expert is selected and updated (solid line);
during the evaluation, experts' outputs are linearly combined with weights produced by $\vg$.}
\end{figure}

Recent efforts by
\citet{voita2019voita} and \citet{michel2019are}
suggest that typical transformer networks are 
overparameterized, in the sense
that at test time, many of the heads, or even a full layer~\citep{fan2019reducing},
 can be removed
without significant loss in performance.\footnote{We do not argue that overparameterization is bad for training.
	In fact, it may be necessary for successful optimization and good generalization~\interalia{neyshabur2014search,zhang2016understanding,soudry2016bad}.
	Rather, we try to explore more efficient ways to use the
	modeling capacity, than, e.g., removing part of the model.} 
In response to this observation, they propose to prune the unimportant attention heads
in the model after it is trained, aiming for faster inference.

In this paper, we ask whether, instead of reducing the model capacity,
we can use it more effectively.
We propose {\bf m}ixture of {\bf a}ttentive {\bf e}xperts (\ourmodel).
\ourmodel retains all attention heads, and learns to activate different heads on different inputs (see illustration in \figref{model}).
We start by showing that multi-head attention can be seen as 
an uniform, input-agnostic mixture of experts \citep{jacobs1991adaptive}, 
by grouping a subset of attention heads as an expert (\S\ref{subsec:multihead_moe}).
We then introduce \ourmodel, which instead of uniformly weighting the experts,
complements the experts with a learned,
input-dependent function that assigns their responsibilities~(\S\ref{subsect:mae}).
To train \ourmodel, we propose a two-step
algorithm based on block coordinate descent~(\S\ref{sec:model}), which alternates between updating the experts' responsibilities 
and their parameters.

We evaluate \ourmodel~on machine translation and language modeling (\S\ref{sec:experiments}).
Our approach outperforms strong baselines on both;
on the WMT14 English to German MT dataset, \ourmodel outperforms transformer-base \citep{vaswani2017attention} by 0.8 BLEU with a negligible increase in the number parameters.
Our analysis shows that \ourmodel learns to encourage different
experts to specialize on different inputs~(\S\ref{sec:analysis}).

	\section{MAE: Mixture of Attentive Experts}
\label{sec:background}
This section describes \ourmodel in detail.
It is inspired by a mixture-of-experts view of multi-head attention,
which we present in \S\ref{subsec:multihead_moe}.
Specifically, we show that multi-head attention
can be viewed as a mixture of uniformly weighted experts,
each consisting of a subset of attention heads.
Based on this observation, we propose \ourmodel,
which learns to weight the experts~(\S\ref{subsect:mae}) depending on the input.
We begin by laying out notation and necessary background in \S\ref{subsec:moe}.

\subsection{Background: Mixture of Experts}\label{subsec:moe}
Mixture of experts is a well-established technique
for ensemble learning~\citep{jacobs1991adaptive}.
It jointly trains a set of \term{expert models} $\{\vf_i\}_{i=1}^{k}$ that 
are intended to specialize across different input cases.
The outputs produced by the experts are aggregated by a linear combination,
with a ``gating function'' $\vg = [g_1,\dots, g_k]$ 
determining the importance of each expert in the final decision:
\begin{align}\label{eq:moe}
\moe(\vx) = \sum_{i=1}^{k} g_i(\vx)\cdot\vf_i(\vx).
\end{align} 
The gating function can be parameterized by, e.g., a neural network.
We will also refer to $\vg$ as the \term{responsibilities} or \term{weights} of the experts.

\subsection{Multi-Head Attention:  a~Mixture-of-Experts~Perspective}
\label{subsec:multihead_moe}
Multi-head attention is the key building block 
for the state-of-the-art transformer architectures~\citep{vaswani2017attention}.
At its core are multiple separately parameterized attention heads.
An attention head takes as input a $n$-by-$d$ matrix $\mX$,
with each row being the vector representation of an input element.
It contextualizes the input using a dot-product attention mechanism:
\begin{align}
\widetilde{\mH}_i
=\softmax\left(\mX\mQ_i\mK_i^\top \mX^\top\right)\mX\mV_i,
\end{align}
where $\mQ_i$, $\mK_i$, and $\mV_i$ are learned matrices,\footnote{Some authors explicitly distinguish
	queries, keys, and values~\citep{vaswani2017attention}.
	These inputs can sometimes differ, e.g.,
	in encoder-decoder attention.
	We suppress such differences for clarity.
}
and the $\softmax$ normalizes row-wise.
The outputs of attention heads are then
concatenated and fed through a learned affine transformation:
\begin{align}\label{eq:multihead:concat}
\mZ \triangleq \multihead\left(\mX\right) =
\left[\widetilde{\mH}_1; \dots;\widetilde{\mH}_h\right]
\mW 
\end{align}
where $\mW$ is a learned matrix,
and $h$ denotes the number of attention heads. 

We now present a different computation equivalent to Eq.~\ref{eq:multihead:concat}, 
aiming for a smoother transition
into following sections. Let $\mH_i = \widetilde{\mH}_i\mW_i$,
where $\mW_i$ is a block submatrix of $\mW$, i.e.,
$\mW=[\mW_1^\top;\mW_2^\top,\dots;\mW_h^\top]^\top$.
Then 
\begin{align}\label{eq:multihead}
	\mZ = \left[\widetilde{\mH}_1; \dots;\widetilde{\mH}_h\right]
	\mW
	=\sum_{i=1}^{h}
	\mH_i.
\end{align}
Eq.~\ref{eq:multihead} provides a different view of the output computation
of the multi-head attention:
each attention head first projects the contextualized representation
with a learned matrix (i.e., $\mH_i = \widetilde{\mH}_i\mW_i$),
then their outputs are gathered with a sum (Eq.~\ref{eq:multihead}).
We now show that this can be seen as a uniformly weighted mixture of experts.
\paragraph{A mixture-of-experts perspective.}
Let us take a closer look at Eq.~\ref{eq:multihead}
and rewrite it:

\begin{align}\label{eq:multihead_moe}
\begin{split}
\mZ&= \frac{1}{h-1}
\sum_{i=1}^h (-1 + h) \; \mH_i \\ 
&=\frac{1}{h-1}
\left(-\sum_{i=1}^h\mH_i + \sum_{i=1}^h\sum_{j=1}^{h}\mH_j\right)\\
&=\sum_{i=1}^h \underbrace{
	\vphantom{ \sum_{j=1}^{h} }
	\frac{1}{h}}_\text{gate $g_i$}
\underbrace{\frac{h}{h-1}\left(- \mH_i+\sum_{j=1}^{h}\mH_j \right)}_\text{expert $\vf_i\left(\mX;\bm{\theta}_i\right)$}.
\end{split}
\end{align}
Eq.~\ref{eq:multihead_moe}
interprets multi-head attention as a mixture of ${h \choose h-1}=h$ experts.
It first constructs a set of $h$ experts $\{\vf_i(\boldsymbol{\cdot};\bm{\theta}_i)\}$,
with $\bm{\theta}_i$ denoting $\vf_i$'s parameters.
$\vf_i(\boldsymbol{\cdot};\bm{\theta}_i)$ 
is a parameterized function of the input,
which calculates a sum of the outputs by
all but the $i$th attention head.
This is achieved by subtracting $\mH_i$
from $\sum_{j=1}^{h}\mH_j$, then scaling up the results by $h/(h-1)$.
The experts share part of the parameters:
any two share $h-2$ attention heads.
A uniform responsibility of $1/h$ is used.

\paragraph{Discussion.}
Viewing multi-head attention through this MoE lens suggests some
interesting consequences.
One can replace the input-agnostic responsibility 
in Eq.~\ref{eq:multihead_moe} with a function over the input.
Indeed, we have good reasons for doing so.
\citet{voita2019voita} and \citet{michel2019are}
show that for transformer networks,
a handful of important attention heads are sufficient to
achieve good test-time performance.
They propose to prune the rest using an input-agnostic procedure.
Instead of doing so,
here we see a potential alternative:
keep all the heads,
but only activate those that are important to the input.
This motivates~\ourmodel, which we now
introduce.

\subsection{\ourmodel: Learning to Weight Experts}
\label{subsect:mae}
\ourmodel is inspired by the connections between MoE and multi-head
attention we draw in \S\ref{subsec:multihead_moe}.
On top of multi-head attention,
\ourmodel learns an input-dependent parameterized gating function $\vg(\boldsymbol{\cdot};\bm{\phi})$ to
complement the experts.
More formally, the uniform responsibility $1/h$ in Eq.~\ref{eq:multihead_moe}
is replaced by $\vg(\boldsymbol{\cdot};\bm{\phi})$:
given input $\mX$, \ourmodel outputs
\begin{align}\label{eq:multihead_moe2}
\sum_{i=1}^{h}
g_i(\mX;\bm{\phi})\cdot \vf_i(\mX; \bm{\theta}_i).
\end{align}
Experts $\vf_i$ are the same as those in Eq.~\ref{eq:multihead_moe}.

$\vg(\boldsymbol{\cdot};\bm{\phi})$ is parameterized with a multi-layer perceptron (MLP)
followed by a $\softmax$.
It first averages $\mX$ along the row 
(i.e., the sequence direction),
and then feeds the results through a two-layer $\tanh$-MLP.
$\vg(\boldsymbol{\cdot};\bm{\phi})$ outputs a normalized $h$-dimensional vector
using a $\softmax$, indicating the responsibilities of the experts.
It can be seen as a learned
probability distribution over the experts.

\ourmodel can learn to assign more responsibility
to the experts that are more important to the given input,
allowing them to contribute more.
\ourmodel~is applicable
wherever multi-head attention is used. 
For example, in a machine translation experiment (\S\ref{subsec:mt}),
we replace with \ourmodel all the multi-head attention in a transformer network,
including the self-attention in all encoder and decoder layers,
as well as those attending over the encoded source from the decoder. 
Each of them is separately treated as a mixture of experts,
and has its own gating function.
The additional parameter overhead is small:
gating functions
account for only 3--5\% parameters of the full model (Appendix~\ref{appendix:implementation}).

	\section{Training \ourmodel with Block Coordinate Descent}
\label{sec:model}

It is straightforward to
jointly train the experts and the gating functions in an \ourmodel model
using backpropagation.
However, in line with previous observations \cite{shen2019mixture}, 
we empirically observe that 
this is prone to degenerate solutions where
the gating functions tend to learn to 
similarly weight the experts (see \S\ref{subsec:cluster}).\footnote{Besides the undesired degeneracy,
we also find that the model suffers worse overfitting when
$\bm{\theta}$ and $\bm{\phi}$ are jointly updated~(Appendix~\ref{subsec:overfit}).
One possible reason is that,
compared to the standard multi-head attention,
the learned gates give the model
additional capacity to compensate for the experts' errors
with others' outputs at \emph{training} time,
hurting generalization~\citep{jacobs1991adaptive}.
Another common degeneracy of MoEs is the ``rich get richer'' where
	one of the experts is always picked and others ignored.
	As observed by~\citet{voita2019voita},
	this can happen when the experts are trained to be
	sparsely weighted.
	When tuning the hyperparameters,
	we observe the ``rich get richer'' degeneracy
	if the learning rate is 
	set too large. \label{fn:overfit}
}

As a remedy, 
we propose a block coordinate descent (BCD) training.
At a high level, training is decomposed into two interleaving steps:
A \term{G step} updates the \text{g}ating function $\vg(\boldsymbol{\cdot};\bm{\phi})$, fixing the experts;
an \term{F step} fixes the gating function and updates \emph{one} randomly selected expert $\vf_i(\boldsymbol{\cdot};\bm{\theta}_i)$.\footnote{For clarity, our discussion focuses on $\bm{\theta}$ and $\bm{\phi}$.
The rest of the model, e.g., the word embeddings in a transformer network,
are updated along with $\bm{\theta}$.
Training aims to minimize loss $\gL$ over $\{\bm{\theta},\bm{\phi}\}$.}
The computations for G and F steps differ:

\begin{compactitem}
	\item In a G step, \ourmodel outputs a linear combination of the experts'
	outputs, and only updates the gating function's parameters~(Algorithm~\ref{algo:g_step}).
	\emph{No} expert is updated.
	\item An F step computes the experts' responsibilities $\vg(\mX)$,
	according to which an expert $i$ is then sampled~(Algorithm~\ref{algo:f_step}). 
	\ourmodel computes the output with  $\vf_i$,
	which is then updated, without updating the gating function or
        other experts.\footnote{In mini-batch training, which we use
          in the experiments, different experts can be sampled for
          different instances in a mini-batch. This is because $\vg$
          depends on the inputs. This means that multiple experts
          will be updated in an F step, but each due to a subset of the examples in
        the mini-batch.}
\end{compactitem}
A non-differentiable sampling from $\vg$
is involved in F steps.
It does not create difficulties for the backpropagation,
since an F step never calculates the gradients w.r.t.~$\bm{\phi}$.
At test time, the computation is the same as that in a G step,
i.e., \ourmodel outputs a linear combination of the experts, weighted by $\vg$.

\paragraph{Training time overhead.}
A straightforward training procedure is to, for each training instance,
first take a G step, and then an F step.
This doubles the forward propagation computation overhead.
In practice,
it is not necessary to take G steps as frequently as F steps,
since they only update a small portion of the model.
In the experiments,
we take G steps one fifth as frequently as F steps:
we make G updates every 5 epochs
while always take F steps.
In preliminary experiments, we find 
this reduces training time overhead without significant
impact on the performance.\footnote{In this way, training time for \ourmodel 
	is roughly 1.2 times longer than that of the transformer network it builds on.}

Algorithm~\ref{algo:bcd} summarizes 
the block coordinate descent training in a given epoch.

\begin{algorithm}[t]
	\centering
	\caption{A G step update for \ourmodel, with step size $\eta$.}
	\label{algo:g_step}
	\begin{algorithmic}[1]
		\Procedure{MaeG}{$\mX$}
		\State $\mZ\leftarrow\sum_{i=1}^{h} 
		g_i(\mX;\bm{\phi})\cdot \vf_i(\mX;\bm{\theta}_i)$
		\State Forwardprop with $\mZ$ and calculate $\gL$. 
		\State Calculate $\nabla_{\bm{\phi}}\gL$ with backprop.
		\State $\bm{\phi}\leftarrow \bm{\phi} - \eta\cdot \nabla_{\bm{\phi}}\gL$.
		
		\EndProcedure
	\end{algorithmic}
\end{algorithm}

\begin{algorithm}[t]
	\centering
	\caption{An F step update for \ourmodel, with step size $\eta$.}
	\label{algo:f_step}
	\begin{algorithmic}[1]
		\Procedure{MaeF}{$\mX$}
		
		\State Draw $i\sim \operatorname{Cat}(\vg(\mX; \bm{\phi}))$
		\State $\mZ\leftarrow\vf_i(\mX;\bm{\theta}_i)$
		\State Forwardprop with $\mZ$ and calculate $\gL$. 
		\State Calculate $\nabla_{\bm{\theta}_i}\gL$ with backprop.
		\State $\bm{\theta}_i\leftarrow \bm{\theta}_i - \eta\cdot \nabla_{\bm{\theta}_i}\gL$.
		\EndProcedure
	\end{algorithmic}
\end{algorithm}

\begin{algorithm}[h]
	\centering
	\caption{Block coordinate descent (BCD) training for \ourmodel, at epoch $e$. $\gD$ denotes the training data.\footnotemark}
	\label{algo:bcd}
	\begin{algorithmic}[1]
		\Procedure{Bcd}{$\gD=\{\mX_i\}_i$, $e$}
		\For{$\mX_i\in \gD$}
		
		\State\Comment Take G steps every 5 epochs.
		\If{$e\bmod 5 = 0$}	
		\State \textsc{MaeG}($\mX_i$)
		\EndIf
		\State\Comment Always do F step updates.
		\State \textsc{MaeF}($\mX_i$)
		\EndFor
		\EndProcedure
	\end{algorithmic}
\end{algorithm}

\footnotetext{Although we assume supervised learning, we suppress 
	the gold outputs for notational clarity.
	We slightly overload the notation and denote by $\mX_i$ the training instance,
	although they cab also be the outputs of intermediate layers. }

\paragraph{Connections to dropout.}
In the above block coordinate descent training algorithm,
an F step samples an expert to update,
and ignores the rest in both forward and backward computation.
It is reminiscent of dropout \citep{srivastava2014dropout}.
Specifically,
selecting expert $\vf_i$ 
is equivalent to dropping head $i$.\footnote{Recall from Eq.~\ref{eq:multihead_moe} that $\vf_i$ includes all but head $i$.}
In other words,
the F steps~(Algorithm~\ref{algo:f_step}) can be seen
as a structured dropout applied to the attention heads,
but with learned input-dependent drop probabilities.
When $\vg$ is a constant vector with elements $1/h$,
it recovers the head dropout,
which is also explored by  concurrent
work~\citep{fan2019reducing}. 

So far, we view \ourmodel as a mixture of $h$ experts,
each consisting of $h-1$ attention heads.
One can, of course, generalize this to other settings, e.g.,
mixing ${h\choose h-2}$ experts, each containing 
$h-2$ heads.
From the dropout view,
this translates to dropping more attention heads:
dropping $t$ heads out of $h$ is equivalent  to applying
a dropout with drop probability $t/h$, in the sense 
that their expected numbers of dropped units
are the same.
  
Despite the similarity between \ourmodel and dropout, a key difference exists between the two:
with the latter, the constant dropout probability is set \emph{a
  priori}, while
\ourmodel uses a gating function $\vg(\boldsymbol{\cdot}; \bm{\phi})$
to calculate a learned,  input-dependent dropout probability.

	\section{Experiments}
\label{sec:experiments}
We empirically evaluate \ourmodel~on machine translation (\S\ref{subsec:mt})
and language modeling (\S\ref{subsec:lm}) benchmarks.
We first introduce the compared models~(\S\ref{subsec:baselines}).

\subsection{Compared Models}\label{subsec:baselines}
\ourmodel is evaluated under two settings:
\begin{compactitem}
\item\ourmodelsc~mixes 8 experts each with 7 attention heads.
\item\ourmodeltsc~is similar to \ourmodelsc, but mixes ${8 \choose 2} = 28$ experts each with 6 attention heads.\footnote{Preliminary results show that mixing experts with fewer heads leads to underwhelming performance. 
	We conjecture this is due to too strong a regularization effect (\S\ref{sec:model}).
}
\end{compactitem}
We compare \ourmodel to the following baselines.
\begin{compactitem}
	\item \base
	is a sequence-to-sequence model based on the transformer architecture.
	\item \joint is the same model as MAE, but
	does not use block coordinate descent training.
	Instead, it jointly updates \emph{all} experts and the gating function at training time, 
	as discussed at the start of \S\ref{sec:model}.

	\item \headdrop  is similar to MAE but
        does not have parameterized gating functions.
	It builds on \base,
	and mixes 8 experts, each with 7 attention heads.
	Constant uniform responsibilities are assigned to the experts.
	At each training step,
	it updates \emph{one} uniformly sampled expert;
	at test time, the outputs of all experts are averaged according to Eq.~\ref{eq:multihead_moe}. 
	\item \headdropt mixes 28 6-attention-head experts, and is otherwise the same as \headdrop.
\end{compactitem}

We refer the readers 
to Appendix~\ref{appendix:implementation} for implementation details.

\subsection{Machine Translation}
\label{subsec:mt}

\paragraph{Datasets.}
We experiment with two machine translation datasets:
\begin{compactitem}
	\item WMT14 EN-DE~\citep{bojar2014wmt}.\footnote{\url{https://drive.google.com/a/haopeng.name/uc?export=download&id=0B_bZck-ksdkpM25jRUN2X2UxMm8}} 
	Following previous practice~\citep{vaswani2017attention} 
	we train on WMT14,
	and designate newstest2013 and newstest2014 as development and test data respectively.
	Our preprocessing follows that of
	\citet{vaswani2017attention} and \citet{ott2018scaling}.
	A shared source-target vocabulary is used,
	with 32k byte pair encoding types~(BPE;~\citealp{sennrich2016bpe}).
	\item IWSLT14 DE-EN~\citep{cettolo2014report}.\footnote{\url{http://workshop2014.iwslt.org/}.}
	It is based on TED talks, and is much smaller compared to WMT14.
	We use the preprocessing from~\citet{edunov2018classical}.
	Following previous practice,
	we use separate vocabularies for the source and target,
	with around 9K and 7K BPE types respectively.
	
\end{compactitem}
Table~\ref{tab:mt_data} summarizes some statistics of the datasets.
\begin{table}[tb]
	\setlength{\tabcolsep}{5pt}
	\centering
	\begin{tabulary}{0.47\textwidth}{@{}l  cccc@{}} 
		
		\toprule
		
		\textbf{Data} & \textbf{Train} & \textbf{Dev.} & \textbf{Test} & \textbf{Vocab.}\\ 
		
		\midrule
		
		WMT14 & 4.5M & 3K & 3K & \phantom{/K}32K\\
		IWSLT14 & 160K& 7K& 7K& 9K/7K\\
		
		\bottomrule
		
	\end{tabulary}
	\caption{Some statistics for WMT14 and IWSLT14 datasets.
		We use separate source and target vocabularies in IWSLT14 experiments.}
	\label{tab:mt_data}
\end{table}

\paragraph{Evaluation.}
The models are evaluated using BLEU~\citep{papineni2002bleu}.
A beam search with beam size 5 is used.
In the WMT14 experiments,
we follow~\citet{vaswani2017attention},
and apply a compound split postprocessing.\footnote{\url{https://github.com/tensorflow/tensor2tensor/blob/master/tensor2tensor/utils/get_ende_bleu.sh}}

\paragraph{Results.}

Table~\ref{tab:res:wmt} summarizes WMT14 EN-DE
 translation test performance.
The base and large sized transformer models are due to~\citet{vaswani2017attention}.
To control for compounding factors,
we additionally compare to our implementation of the base sized model (\base).
It achieves slightly better performance than~\citet{vaswani2017attention},
with a 0.3 BLEU edge.
\ourmodelsc~improves over the base transformer by 0.8 BLEU, obtaining similar performance to the
large-size transformer of~\citet{vaswani2017attention}
using less than a third as many parameters. 
Since
we do not see similar improvement by
\headdrop, we attribute this gain to input-dependent
expert weighting. 
Having a smaller number of heads for each expert,
\ourmodeltsc slightly underperforms \ourmodelsc, 
and so does \headdropt in comparison to \headdrop.
Finally, \joint gets worse performance than the transformer baseline, demonstrating the importance of the block coordinate decent training.

We observe similar trends on the IWSLT14 DE-EN
dataset, summarized in Table~\ref{tab:res:iwslt}.
The \base model here is similar to the base-sized transformer in the WMT14 experiment,
but with a smaller hidden dimension.
\ourmodelsc outperforms \base by 0.9 BLEU.
Interestingly, \headdrop improves over \base by 0.3 BLEU,
possibly because the regularization effect of random expert selection training
helps more on this smaller dataset.\footnote{Selecting an expert
can be seen dropping one attention head in training~(\S\ref{sec:model}).}

\begin{table}[tb]
	\centering
	\begin{tabulary}{0.47\textwidth}{@{}l  c@{\hskip 0.25cm}c@{}} 
		
		\toprule
		\textbf{Model}
		& \textbf{BLEU}		& \textbf{\# Params.}\\
		
		\midrule[.03em]
	
		Base Transformer & 27.3 & 65M\\
		Large Transformer & 28.4 & 213M\\
		\base & 27.6 & 61M\\
		
		\midrule[.03em]
		$^\ddagger$\joint & 27.5 & 63M\\ 
		$^{\dagger}$\headdrop & 27.7 & 61M\\ 
		$^{\dagger}$\headdropt & 27.6 & 61M\\

		\midrule[.03em]

		$^{\dagger\ddagger}$\ourmodelsc & \bf{28.4} & 63M\\ 
		$^{\dagger\ddagger}$\ourmodeltsc & 28.1 & 63M\\ 
		
		\bottomrule
		
	\end{tabulary}
	\caption{WMT14 EN-DE translation test performance on newstest2014.
		$\dagger$ randomly select an expert to update for each training instance,
		and $\ddagger$ learns a gating function to weight the
                experts.
		Transformer performance in the first two rows are due
                to~\citet{vaswani2017attention}.
		}
	\label{tab:res:wmt}
\end{table}

\begin{table}[tb]
	\centering
	\begin{tabulary}{0.47\textwidth}{@{}l  c@{\hskip 0.25cm}c@{}} 
		
		\toprule
		\textbf{Model}
		& \textbf{BLEU}
		& \textbf{\# Params.}\\
		
		\midrule[.03em]
		
		\base 
		& 34.6 & 39M \\
		
		\midrule[.03em]
		$^\ddagger$\joint & 34.8 & 41M\\
		$^{\dagger}$\headdrop & 34.9 & 39M\\
		$^{\dagger}$\headdropt & 35.0 & 39M\\

		\midrule[.03em]

		$^{\dagger\ddagger}$\ourmodelsc & \textbf{35.5} & 41M\\
		$^{\dagger\ddagger}$\ourmodeltsc & 35.4 & 41M\\
		
		\bottomrule
		
	\end{tabulary}
	\caption{IWSLT14 GE-DE test set performance.
		See Table~\ref{tab:res:wmt} caption for indications of the superscripts.}
	\label{tab:res:iwslt}
\end{table}

\subsection{Token-level Language Modeling}\label{subsec:lm}
\paragraph{Dataset.}
We experiment with the WikiText-103 dataset~\citep{merity2016pointer}.
It contains articles from English Wikipedia,
with a 268K-sized vocabulary.
The training/development/test data respectively have
103M/218K/246K tokens.

\paragraph{Setting.}
Here the \base model is 
the strong language model by~\citet{baevski2018adaptive}.
It is based on a 16-layer transformer network; 
each multi-head attention layer has 8 heads.
It uses different embedding dimensions for the tokens,
based on their frequencies.
We closely follow \citet{baevski2018adaptive} in terms of 
hyperparameters and training procedures.
The readers are referred to their paper and Appendix~\ref{appendix:implementation}
for further architecture and hyperparameter details.

\paragraph{Notes on context size.}
\citet{baevski2018adaptive} study the effect of context window,
i.e., the number of history tokens the model attends over.
They find that using larger context sizes 
lead to better performance~(\citealp{baevski2018adaptive}, Table 5).
Their best setting uses a 3,072 training context size, and
2,048 at test time (i.e., the model has access 2,048 tokens before predicting any token at test time).
However, we are not able to train~\ourmodel, nor replicate their results,
 under this setting---our 
GPUs have far less memory, and it is impossible
to even load a 3,072-token 
context chunk.\footnote{\citet{baevski2018adaptive} use NVIDIA Tesla V100 GPUs with 32GB memory,
	while we only have access to GeForce RTX 2080 Ti, with 11GB memory.
	}
Therefore we train and evaluate \ourmodel and \headdrop
with smaller 512/480 context sizes, also explored by
	\citet{baevski2018adaptive},
	which allows for a head-to-head comparison.

\paragraph{Results.}
Table~\ref{tab:res:wikitext} shows the perplexity on WikiText-103 test data.
When trained under the same setting,
\ourmodel outperforms \citet{baevski2018adaptive} by more than 0.3 perplexity.
Interestingly, despite the much smaller context at both training and test time,
\ourmodel matches the best setting by \citet{baevski2018adaptive}.
\headdrop and \joint \emph{underperform} the baseline (higher perplexity).

\begin{table}[tb]
	\centering
	\begin{tabulary}{0.47\textwidth}{@{}l  c@{\hskip 0.25cm}c@{}} 
		
		\toprule
		\textbf{Model}
		& \textbf{Perplexity}
		& \textbf{\# Params.}\\
		
		\midrule[.03em]
		
		$^\star$\base (B\&A, 2019) & 18.70 & 247M \\
		
		\midrule[.03em]

		\base (B\&A, 2019) & 19.03 & 247M \\
		\midrule[.03em]
		
		$^\ddagger$\joint & 19.12 & 249M\\
		$^{\dagger}$\headdrop & 19.26 & 247M\\
		\midrule[.03em]
		$^{\dagger\ddagger}$\ourmodelsc & \textbf{18.71}  & 249M\\
		
		\bottomrule
		
	\end{tabulary}
	\caption{Language modeling performance on WikiText-103 test set (lower is better).
		$\star$Trains/evaluates with 3,072/2,048 context
                sizes and therefore not directly comparable
                to other models which use 512/480 sized ones.
		See Table~\ref{tab:res:wmt} caption for the indications of other superscripts.
		Bold font indicates the best performance using smaller context sizes.
		The first two rows are due to Table 5 of \citet{baevski2018adaptive}.
		} 
	\label{tab:res:wikitext}
\end{table}
	\section{Analysis}
\label{sec:analysis}
This section first empirically confirms that \ourmodel learns to activate different experts
on different inputs in \S\ref{subsec:cluster}.
We then run a synthetic experiment to explore
\ourmodel's potential in transfer learning~(\S\ref{subsec:finetune}).

\subsection{Does MAE Learn to Specialize the Experts?}\label{subsec:cluster}
One of the appealing properties of MoE models is that
they could 
learn to activate different experts,
depending on what ``expertise'' is needed for the input.
Does \ourmodel learn to do so? 
We empirically study this question, and present evidence indicating that it does,
at least in part.
We consider the encoders of the \headdrop, \joint, and the \ourmodelsc models trained on WMT14.\footnote{The same experiments can be done with the decoders,
	where the inputs to gating functions are German sentences.
	The authors lack German expertise, and interpretation of a following
analysis would not have been possible for us.
}

\begin{table}[tb]
	\centering
	\begin{tabulary}{0.47\textwidth}{@{}l @{\hskip 0.4cm} l@{\hskip 0.7cm} l @{}} 
		
		\toprule
		\textbf{Model}
		& \textbf{BLEU}
		& \textbf{Diff.}\\
		
		\midrule[.03em]
		
		\headdrop    & 26.6 & - \\ 
		\quad One random expert    & 25.8$_{\pm 0.2}$ & $\downarrow$ 0.8$_{\pm 0.2}$ \\ 
		\midrule[.03em]
		\joint   & 26.7 & - \\ 
		\quad Most specialized expert    & 26.0 & $\downarrow$ 0.7  \\ 
		\midrule[.03em]
		\ourmodelsc  & 27.1 &  -\\ 
		\quad Most specialized expert    & 26.8 & $\downarrow$ 0.3\\ 
		
		\bottomrule
		
	\end{tabulary}
	\caption{Performance decrease for different models on WMT14 development set when only one expert is used for each multi-head attention layer~(\ref{subsec:cluster}). }
	\label{tab:specialize}
\end{table}

We first study whether BCD training helps drifting \ourmodel away from
uniformly weighting the experts agnostic to the inputs.
We treat the gating values as probabilities, and 
calculate their entropies: $\mathcal{H}(\vg) = -\sum_{i=1}^{h}g_i\cdot\log g_i$,
which are then averaged across different layers.
The average entropy on the development set for \ourmodelsc 
is 1.91, lower than the 2.02 by the \joint model trained without BCD.
In comparison, \headdrop uniformly weights the experts and has the
entropy of 2.08.
This indicates that gating weights of \ourmodel trained with 
BCD are more ``focused'' on one or a subset of experts than trained
without.

Second, we study whether \ourmodel learns to specialize different
experts for different inputs. 
To do so we attribute the development instances to the experts
that maximize the gating weights.
For the first encoder layer of \ourmodelsc,
the percentages of instances attributed to each of the 8 experts are 
relatively balanced:
13\%, 14\%, 9\%, 16\%, 10\%, 15\%, 10\%, 12\%.\footnote{We observe similar trends in other layers. See Appendix~\ref{subsec:attribute} for more details.}
This suggests 
that all experts are assigned a substantial part of the input, and it is \emph{not} the case that BCD leads to a ``rich get richer'' outcome.

We then continue and explore whether \ourmodel performs reasonably well when using only the most ``specialized'' experts.
For each development instance,
we select those experts maximizing the gating weights
and ignore the rest,
instead of linearly combining them as in Eq.~\ref{eq:multihead_moe2}.
We see from
Table~\ref{tab:specialize} a 0.3 BLEU decrease under this setting.
In comparison, \joint has a larger performance decrease of 0.7 BLEU.
\joint's performance drop is similar to that of \headdrop,
for which we randomly select an expert at each layer and average the 
performance over 5 runs.
These results support the proposition that \ourmodel specializes better
when trained with BCD.

Finally, we search for the tokens that are more likely to activate each expert.
We compute the pointwise mutual information (PMI; \citealp{church1990word}) between
 tokens and experts:
\begin{align*}
\operatorname{PMI}(\textrm{token}_i, \textrm{expert}_j)=
\log\frac{p(\textrm{token}_i, \textrm{expert}_j)}{p(\textrm{token}_i)p(\textrm{expert}_j)}.
\end{align*}
Table~\ref{tab:token_analysis}
lists the most indicative tokens of each expert, for the first layer.
While some of the terms for some experts seem loosely related (e.g.,
\emph{bell, reuters}, and {\it computing} for expert 2, it is hard to find clear patterns in most of them.

\begin{table}[t]
	\centering
	\small
	\begin{tabulary}{\columnwidth}{@{}c c c c@{}}
		\toprule
		
		Expert 1 & Expert 2 & Expert 3 & Expert 4 \\
		\midrule[.03em]
		neumann & bell & candidacy & veil\\
		debuted & zero & rose & monument\\
		rental & computing & submission & fox\\
		worthy & decentralized & palm  & unnerved \\
		landloards & reuters & roles & remainder\\
		\midrule[.15em]
		
		Expert 5 & Expert 6 & Expert 7 & Expert 8 \\
		\midrule[.03em]
		
		spoil & menses & romans & odds\\
		anybody & technological & sticker & heat\\
		endorsed & inevitably & outdated   & marvel\\
		reserve &bet & analyst  & ornate\\
		pending & punk & venues  & anticipating\\
		
		\bottomrule
	\end{tabulary}
	\caption{Indicative tokens for each expert~(\S\ref{subsec:cluster}).
		Tokens attributed to Expert 2
		are mostly computer science terminology;
		trends for other experts are less clear.
	}
	\label{tab:token_analysis}
\end{table}

\subsection{\ourmodel's Potential in Transfer Learning: A~Case~Study}\label{subsec:finetune}
We now turn to evaluate another property of \ourmodel: 
its potential for data-efficient transfer learning,
by only updating the gating functions, freezing the experts.
We consider the pretrain-then-finetune setting.
Due to computation limits, we are unable to
explore \ourmodel for pre-training contextual representations~\cite{peters2018deep,delvin2019bert}.
Rather, we focus on the following small-scale 
machine translation experiments.

\paragraph{Setting.}
We explore finetuning on IWSLT14 EN-DE data, a \ourmodel model pretrained 
on the much larger WMT14 dataset.\footnote{Here we reverse 
	the translation direction of IWSLT14:
	\S\ref{subsec:mt} experimented with DE-EN,
	here we use EN-DE.
	}
We compare three finetuning methods:
\begin{compactitem}
	\item \finetuneg finetunes the gating functions' parameters (i.e., $\bm{\phi}$),
	keeping the rest frozen.
	\item \finetunegp updates the parameter matrix $\mW$ in Eq.~\ref{eq:multihead}
	in addition to $\bm{\phi}$. The rest of the model parameters are fixed.
	\item \finetuneall updates all parameters.
\end{compactitem}
As a baseline, \nofinetune is the out-of-box pretrained model without any finetuning.
\scratch  trains a \ourmodel model from scratch.

Table~\ref{tab:res:transfer} summarizes
the IWSLT14 EN-DE development set performance.
Surprisingly, 
\nofinetune already outperforms \scratch without any finetuning.
We attribute this improvement to the larger pretraining (WMT14) data.
Only updating the gating functions,
\finetuneg improves over \nofinetune
by 0.8 BLEU.
Yet there is still a significant gap of 1.8 BLEU between 
\finetuneg and \finetuneall.
Interestingly,  \finetunegp almost matches 
the performance of \finetuneall, but only updates
1/9 as many parameters.
Both \finetuneg and \finetunegp reach the best 
performance after around 1K gradient updates, i.e., one epoch,
significantly less than \finetuneall or \scratch.
	
\begin{figure}
	\centering
	\includegraphics[clip,trim=.0cm 0cm 0cm 0.9cm,width=.7\columnwidth]{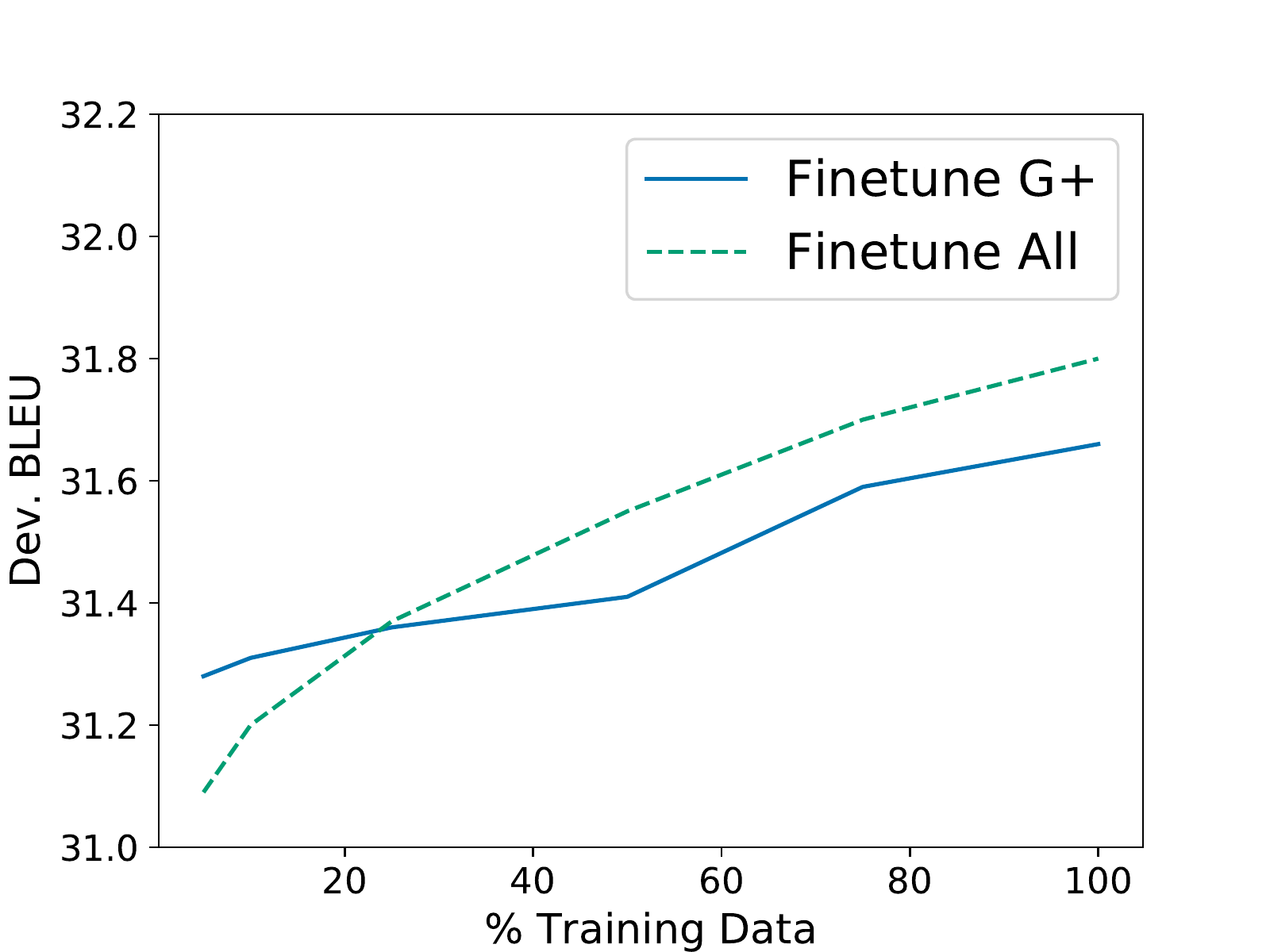}
	\caption{IWSLT14 development performance of \finetunegp and \finetuneall
		using different amount of training data~(\S\ref{subsec:finetune}).
		When trained on less than 20\% subset of the original training data,
		\finetunegp outperforms \finetuneall.}
	\label{fig:transfer}
\end{figure}

We further compare \finetunegp and \finetuneall where 
less downstream training data is available.
To simulate this, we randomly sample [5\%, 10\%, 25\%, 50\%, 75\%]
subsets of IWSLT14 training data,
on which the pretrained model is finetuned.
Figure~\ref{fig:transfer} plots their performance.
We see a clear trend: 
as less training data is available,
the gap between \finetunegp and \finetuneall decreases;
when less than 20\% of the training data is available,
\finetunegp outperforms \finetuneall. 
These results suggest that finetuning 
\ourmodel with \finetunegp can be viable in low-resource transfer learning.
 
\begin{table}[tb]
	\centering
	\begin{tabulary}{0.47\textwidth}{@{}l  c@{\hskip 0.25cm}c c@{}} 
		
		\toprule
		\textbf{Method}
		& \textbf{BLEU}
		& \textbf{\# Params.}
		& \textbf{\# Steps.}\\
		
		\midrule[.03em]
		
		\scratch & 28.8 & 41M & 52K\\
		\midrule[.03em]
		\nofinetune & 29.3 & 0& 0\\

		\midrule[.03em]
		\finetuneg & 30.1 & 2M & 1K \\
		\finetunegp & 31.6 & 7M& 1K\\
		\finetuneall & \textbf{31.8} & 63M & 12K\\
		
		\bottomrule
		
	\end{tabulary}
	\caption{IWSLT14 development set performance
		of different finetuning methods~(\S\ref{subsec:finetune}).
		The last two columns indicate the number of parameters to update,
		and the number of gradient steps needed to achieve the best development performance.
		}
	\label{tab:res:transfer}
\end{table}

	\section{Related Work}
\label{sec:related}

\paragraph{Multi-head attention.} 
An increasing amount of effort has been devoted into 
developing better attention mechanisms~\interalia{malaviya2018sparse,kim2018latent,sukhbaatar2019adaptive,correia2019adaptively,maruf2019selective},
and improving transformer architectures ~\interalia{shaw2018self,dehghani2018universal,hao2019modeling,correia2019adaptively,Baosong2019convolutional}.
Closely related, \citet{iida2019attention} applies another attention mechanism over the attention heads, allowing a learned reweighting of them.
Our work focuses on the connection between multi-head attention and
MoE, and the BCD training it suggests and benefits from.
Concurrent to our work, \citep{fan2019reducing} study structurally pruning transformer layers
for more efficient inference.

Another line of work aims to better understand 
the working of transformer models~\interalia{clark2019what,liu2019linguistic,tenney2019bert}.

\paragraph{Mixture of experts.} 
One of the most successful applications of 
MoE is ensemble learning~\interalia{caruana2004ensemble,liu2017robust,dutt2017coupled}.
Recent efforts also explore MoE in
sequence learning~\citep{shazeer2017outrageously},
and to promote diversity in text generation~\interalia{he2018sequence,shen2019mixture,cho2019mixture}. 
	\section{Conclusion}
We presented \ourmodel.
It is inspired by a mixture-of-experts perspective of multi-head attention.
With a learned gating function, \ourmodel activates different experts on different inputs.
\ourmodel is trained using a block coordinate descent algorithm,
 which alternates between updating the responsibilities of the experts and their parameters. 
Our experiments show that \ourmodel outperforms the transformer baselines 
on machine translation and language modeling benchmarks.
The analysis shows that \ourmodel learns to activate different experts.
The code is publicly available at \url{https://github.com/Noahs-ARK/MAE}.

	\section*{Acknowledgments}
	We thank the anonymous reviewers,
	Yoav Artzi, 
	Mandar Joshi, 
	Jungo Kasai,
	Lingpeng Kong, 
	Kenton Lee, 
	Kelvin Luu, 
	Will Merrill, 
	Phoebe Mulcaire, 
	Mark Neumann, 
	Nikos Pappas, 
	Ofir Press,
	Lianhui Qin, 
	Swabha Swayamdipta, 
	Vivek Srikumar,
	Sam Thomson,
	and Dani Yogatama
	for their helpful feedback.
	This work was supported in part by NSF grant 1562364, a Google Fellowship, and 
	NVIDIA Corporation through the donation of a Tesla GPU.

	\bibliography{acl2020}

\begin{thebibliography}{45}
\expandafter\ifx\csname natexlab\endcsname\relax\def\natexlab#1{#1}\fi

\bibitem[{Baevski and Auli(2019)}]{baevski2018adaptive}
Alexei Baevski and Michael Auli. 2019.
\newblock Adaptive input representations for neural language modeling.
\newblock In \emph{Proc. of ICLR}.

\bibitem[{Bojar et~al.(2014)Bojar, Buck, Federmann, Haddow, Koehn, Leveling,
  Monz, Pecina, Post, Saint-Amand, Soricut, Specia, and
  Tamchyna}]{bojar2014wmt}
Ond{\v{r}}ej Bojar, Christian Buck, Christian Federmann, Barry Haddow, Philipp
  Koehn, Johannes Leveling, Christof Monz, Pavel Pecina, Matt Post, Herve
  Saint-Amand, Radu Soricut, Lucia Specia, and Ale{\v{s}} Tamchyna. 2014.
\newblock Findings of the 2014 workshop on statistical machine translation.
\newblock In \emph{Proc. of WMT}.

\bibitem[{Caruana et~al.(2004)Caruana, Niculescu-Mizil, Crew, and
  Ksikes}]{caruana2004ensemble}
Rich Caruana, Alexandru Niculescu-Mizil, Geoff Crew, and Alex Ksikes. 2004.
\newblock Ensemble selection from libraries of models.
\newblock In \emph{Proc. of ICML}.

\bibitem[{Cettolo et~al.(2014)Cettolo, Niehues, Stüker, Bentivogli, and
  Federico}]{cettolo2014report}
Mauro Cettolo, Jan Niehues, Sebastian Stüker, Luisa Bentivogli, and Marcello
  Federico. 2014.
\newblock Report on the 11th {IWSLT} evaluation campaign.
\newblock In \emph{Proc. of IWSLT}.

\bibitem[{Cho et~al.(2019)Cho, Seo, and Hajishirzi}]{cho2019mixture}
Jaemin Cho, Minjoon Seo, and Hannaneh Hajishirzi. 2019.
\newblock Mixture content selection for diverse sequence generation.
\newblock In \emph{Proc. of EMNLP}.

\bibitem[{Church and Hanks(1990)}]{church1990word}
Kenneth~Ward Church and Patrick Hanks. 1990.
\newblock Word association norms, mutual information, and lexicography.
\newblock \emph{Computational Linguistics}, 16(1):22--29.

\bibitem[{Clark et~al.(2019)Clark, Khandelwal, Levy, and
  Manning}]{clark2019what}
Kevin Clark, Urvashi Khandelwal, Omer Levy, and Christopher~D. Manning. 2019.
\newblock What does {BERT} look at? an analysis of {BERT}'s attention.
\newblock In \emph{Proc. of BlackBoxNLP}.

\bibitem[{Correia et~al.(2019)Correia, Niculae, and
  Martins}]{correia2019adaptively}
Gon{\c{c}}alo~M. Correia, Vlad Niculae, and Andr{\'e}~F.T. Martins. 2019.
\newblock Adaptively sparse transformers.
\newblock In \emph{Proc. of EMNLP}.

\bibitem[{Dehghani et~al.(2019)Dehghani, Gouws, Vinyals, Uszkoreit, and
  Kaiser}]{dehghani2018universal}
Mostafa Dehghani, Stephan Gouws, Oriol Vinyals, Jakob Uszkoreit, and {\L}ukasz
  Kaiser. 2019.
\newblock Universal transformers.

\bibitem[{Deng et~al.(2018)Deng, Kim, Chiu, Guo, and Rush}]{kim2018latent}
Yuntian Deng, Yoon Kim, Justin Chiu, Demi Guo, and Alexander Rush. 2018.
\newblock Latent alignment and variational attention.
\newblock In \emph{Proc. of NeurIPS}.

\bibitem[{Devlin et~al.(2019)Devlin, Chang, Lee, and
  Toutanova}]{delvin2019bert}
Jacob Devlin, Ming-Wei Chang, Kenton Lee, and Kristina Toutanova. 2019.
\newblock {BERT}: Pre-training of deep bidirectional transformers for language
  understanding.
\newblock In \emph{Proc. of NAACL}.

\bibitem[{Dutt et~al.(2017)Dutt, Pellerin, and Quénot}]{dutt2017coupled}
Anuvabh Dutt, Denis Pellerin, and Georges Quénot. 2017.
\newblock Coupled ensembles of neural networks.
\newblock {arXiv}:1709.06053.

\bibitem[{Edunov et~al.(2018)Edunov, Ott, Auli, Grangier, and
  Ranzato}]{edunov2018classical}
Sergey Edunov, Myle Ott, Michael Auli, David Grangier, and Marc{'}Aurelio
  Ranzato. 2018.
\newblock Classical structured prediction losses for sequence to sequence
  learning.
\newblock In \emph{Proc. of NAACL}.

\bibitem[{Fan et~al.(2020)Fan, Grave, and Joulin}]{fan2019reducing}
Angela Fan, Edouard Grave, and Armand Joulin. 2020.
\newblock Reducing transformer depth on demand with structured dropout.
\newblock In \emph{Proc. of ICLR}.

\bibitem[{Hao et~al.(2019)Hao, Wang, Yang, Wang, Zhang, and
  Tu}]{hao2019modeling}
Jie Hao, Xing Wang, Baosong Yang, Longyue Wang, Jinfeng Zhang, and Zhaopeng Tu.
  2019.
\newblock Modeling recurrence for transformer.
\newblock In \emph{Proc. of NAACL}.

\bibitem[{He et~al.(2018)He, Haffari, and Norouzi}]{he2018sequence}
Xuanli He, Gholamreza Haffari, and Mohammad Norouzi. 2018.
\newblock Sequence to sequence mixture model for diverse machine translation.
\newblock In \emph{Proc. of CoNLL}.

\bibitem[{Iida et~al.(2019)Iida, Kimura, Cui, Hung, Utsuro, and
  Nagata}]{iida2019attention}
Shohei Iida, Ryuichiro Kimura, Hongyi Cui, Po-Hsuan Hung, Takehito Utsuro, and
  Masaaki Nagata. 2019.
\newblock Attention over heads: A multi-hop attention for neural machine
  translation.
\newblock In \emph{Proc. of ACL: Student Research Workshop}.

\bibitem[{Ioffe and Szegedy(2015)}]{sergey2015batchnorm}
Sergey Ioffe and Christian Szegedy. 2015.
\newblock Batch normalization: Accelerating deep network training by reducing
  internal covariate shift.
\newblock In \emph{Proc. of ICML}.

\bibitem[{Jacobs et~al.(1991)Jacobs, Jordan, Nowlan, and
  Hinton}]{jacobs1991adaptive}
Robert~A. Jacobs, Michael~I. Jordan, Steven~J. Nowlan, and Geoffrey~E. Hinton.
  1991.
\newblock Adaptive mixtures of local experts.
\newblock \emph{Neural Computation}, 3(1):79--87.

\bibitem[{Liu et~al.(2019{\natexlab{a}})Liu, Gardner, Belinkov, Peters, and
  Smith}]{liu2019linguistic}
Nelson~F. Liu, Matt Gardner, Yonatan Belinkov, Matthew~E. Peters, and Noah~A.
  Smith. 2019{\natexlab{a}}.
\newblock Linguistic knowledge and transferability of contextual
  representations.
\newblock In \emph{Proc. of NAACL}.

\bibitem[{Liu et~al.(2018)Liu, Cheng, Zhang, and Hsieh}]{liu2017robust}
Xuanqing Liu, Minhao Cheng, Huan Zhang, and Cho-Jui Hsieh. 2018.
\newblock Towards robust neural networks via random self-ensemble.
\newblock In \emph{Proc. of ECCV}.

\bibitem[{Liu et~al.(2019{\natexlab{b}})Liu, Ott, Goyal, Du, Joshi, Chen, Levy,
  Lewis, Zettlemoyer, and Stoyanov}]{liu2019roberta}
Yinhan Liu, Myle Ott, Naman Goyal, Jingfei Du, Mandar Joshi, Danqi Chen, Omer
  Levy, Mike Lewis, Luke Zettlemoyer, and Veselin Stoyanov. 2019{\natexlab{b}}.
\newblock {RoBERTa}: {A} robustly optimized {BERT} pretraining approach.
\newblock {arXiv}:1907.11692.

\bibitem[{Malaviya et~al.(2018)Malaviya, Ferreira, and
  Martins}]{malaviya2018sparse}
Chaitanya Malaviya, Pedro Ferreira, and Andr{\'e}~FT Martins. 2018.
\newblock Sparse and constrained attention for neural machine translation.
\newblock In \emph{Proc. of ACL}.

\bibitem[{Maruf et~al.(2019)Maruf, Martins, and Haffari}]{maruf2019selective}
Sameen Maruf, Andr{\'e}~FT Martins, and Gholamreza Haffari. 2019.
\newblock Selective attention for context-aware neural machine translation.
\newblock In \emph{Proc. of NAACL}.

\bibitem[{Merity et~al.(2016)Merity, Xiong, Bradbury, and
  Socher}]{merity2016pointer}
Stephen Merity, Caiming Xiong, James Bradbury, and Richard Socher. 2016.
\newblock Pointer sentinel mixture models.
\newblock {arXiv}:1609.07843.

\bibitem[{Michel et~al.(2019)Michel, Levy, and Neubig}]{michel2019are}
Paul Michel, Omer Levy, and Graham Neubig. 2019.
\newblock Are sixteen heads really better than one?
\newblock In \emph{Proc. of NeurIPS}.

\bibitem[{Neyshabur et~al.(2014)Neyshabur, Tomioka, and
  Srebro}]{neyshabur2014search}
Behnam Neyshabur, Ryota Tomioka, and Nathan Srebro. 2014.
\newblock In search of the real inductive bias: On the role of implicit
  regularization in deep learning.
\newblock In \emph{Proc. of ICLR: Worshop Tack}.

\bibitem[{Ott et~al.(2018)Ott, Edunov, Grangier, and Auli}]{ott2018scaling}
Myle Ott, Sergey Edunov, David Grangier, and Michael Auli. 2018.
\newblock Scaling neural machine translation.
\newblock In \emph{Proc. of WMT}.

\bibitem[{Papineni et~al.(2002)Papineni, Roukos, Ward, and jing
  Zhu}]{papineni2002bleu}
Kishore Papineni, Salim Roukos, Todd Ward, and Wei jing Zhu. 2002.
\newblock {BLEU}: a method for automatic evaluation of machine translation.
\newblock In \emph{Proc. of ACL}.

\bibitem[{Peters et~al.(2018)Peters, Neumann, Iyyer, Gardner, Clark, Lee, and
  Zettlemoyer}]{peters2018deep}
Matthew~E. Peters, Mark Neumann, Mohit Iyyer, Matt Gardner, Christopher Clark,
  Kenton Lee, and Luke Zettlemoyer. 2018.
\newblock Deep contextualized word representations.
\newblock In \emph{Proc. of NAACL}.

\bibitem[{Radford et~al.(2018)Radford, Wu, Child, Luan, Amodei, and
  Sutskever}]{radford2018language}
Alec Radford, Jeffrey Wu, Rewon Child, David Luan, Dario Amodei, and Ilya
  Sutskever. 2018.
\newblock Language models are unsupervised multitask learners.

\bibitem[{Sennrich et~al.(2016)Sennrich, Haddow, and Birch}]{sennrich2016bpe}
Rico Sennrich, Barry Haddow, and Alexandra Birch. 2016.
\newblock Neural machine translation of rare words with subword units.
\newblock In \emph{Proc. of ACL}.

\bibitem[{Shaw et~al.(2018)Shaw, Uszkoreit, and Vaswani}]{shaw2018self}
Peter Shaw, Jakob Uszkoreit, and Ashish Vaswani. 2018.
\newblock Self-attention with relative position representations.
\newblock In \emph{Proc. of NAACL}.

\bibitem[{Shazeer et~al.(2017)Shazeer, Mirhoseini, Maziarz, Davis, Le, Hinton,
  and Dean}]{shazeer2017outrageously}
Noam Shazeer, Azalia Mirhoseini, Krzysztof Maziarz, Andy Davis, Quoc Le,
  Geoffrey Hinton, and Jeff Dean. 2017.
\newblock Outrageously large neural networks: The sparsely-gated
  mixture-of-experts layer.
\newblock {arXiv}:1701.06538.

\bibitem[{Shen et~al.(2019)Shen, Ott, Auli, and Ranzato}]{shen2019mixture}
Tianxiao Shen, Myle Ott, Michael Auli, and Marc'Aurelio Ranzato. 2019.
\newblock Mixture models for diverse machine translation: Tricks of the trade.
\newblock In \emph{Proc. of ICML}.

\bibitem[{Soudry and Carmon(2016)}]{soudry2016bad}
Daniel Soudry and Yair Carmon. 2016.
\newblock No bad local minima: Data independent training error guarantees for
  multilayer neural networks.
\newblock {arXiv}:1605.08361.

\bibitem[{Srivastava et~al.(2014)Srivastava, Hinton, Krizhevsky, Sutskever, and
  Salakhutdinov}]{srivastava2014dropout}
Nitish Srivastava, Geoffrey Hinton, Alex Krizhevsky, Ilya Sutskever, and Ruslan
  Salakhutdinov. 2014.
\newblock Dropout: A simple way to prevent neural networks from overfitting.
\newblock \emph{JMLR}, 15(1):1929--1958.

\bibitem[{Strubell et~al.(2018)Strubell, Verga, Andor, Weiss, and
  McCallum}]{strubell2018linguistically}
Emma Strubell, Patrick Verga, Daniel Andor, David Weiss, and Andrew McCallum.
  2018.
\newblock Linguistically-informed self-attention for semantic role labeling.
\newblock In \emph{Proc. of EMNLP}.

\bibitem[{Sukhbaatar et~al.(2019)Sukhbaatar, Grave, Bojanowski, and
  Joulin}]{sukhbaatar2019adaptive}
Sainbayar Sukhbaatar, Edouard Grave, Piotr Bojanowski, and Armand Joulin. 2019.
\newblock Adaptive attention span in transformers.
\newblock In \emph{Proc. of ACL}.

\bibitem[{Tenney et~al.(2019)Tenney, Das, and Pavlick}]{tenney2019bert}
Ian Tenney, Dipanjan Das, and Ellie Pavlick. 2019.
\newblock {BERT} rediscovers the classical {NLP} pipeline.
\newblock In \emph{Proc. of ACL}.

\bibitem[{Vaswani et~al.(2017)Vaswani, Shazeer, Parmar, Uszkoreit, Jones,
  Gomez, Kaiser, and Polosukhin}]{vaswani2017attention}
Ashish Vaswani, Noam Shazeer, Niki Parmar, Jakob Uszkoreit, Llion Jones,
  Aidan~N. Gomez, Lukasz Kaiser, and Illia Polosukhin. 2017.
\newblock Attention is all you need.
\newblock In \emph{Proc. of NeurIPS}.

\bibitem[{Voita et~al.(2019)Voita, Talbot, Moiseev, Sennrich, and
  Titov}]{voita2019voita}
Elena Voita, David Talbot, Fedor Moiseev, Rico Sennrich, and Ivan Titov. 2019.
\newblock Analyzing multi-head self-attention: Specialized heads do the heavy
  lifting, the rest can be pruned.
\newblock In \emph{Proc. of ACL}.

\bibitem[{Yang et~al.(2019{\natexlab{a}})Yang, Wang, Wong, Chao, and
  Tu}]{Baosong2019convolutional}
Baosong Yang, Longyue Wang, Derek~F. Wong, Lidia~S. Chao, and Zhaopeng Tu.
  2019{\natexlab{a}}.
\newblock Convolutional self-attention networks.
\newblock In \emph{Proc. of NAACL}.

\bibitem[{Yang et~al.(2019{\natexlab{b}})Yang, Dai, Yang, Carbonell,
  Salakhutdinov, and Le}]{yang2019xlnet}
Zhilin Yang, Zihang Dai, Yiming Yang, Jaime~G. Carbonell, Ruslan Salakhutdinov,
  and Quoc~V. Le. 2019{\natexlab{b}}.
\newblock {XLNet}: Generalized autoregressive pretraining for language
  understanding.
\newblock {arXiv}:1906.08237.

\bibitem[{Zhang et~al.(2016)Zhang, Bengio, Hardt, Recht, and
  Vinyals}]{zhang2016understanding}
Chiyuan Zhang, Samy Bengio, Moritz Hardt, Benjamin Recht, and Oriol Vinyals.
  2016.
\newblock Understanding deep learning requires rethinking generalization.
\newblock In \emph{Proc. of ICLR}.

\end{thebibliography}
      	\bibliographystyle{acl_natbib}
	\clearpage
	\begin{appendices}
\label{sec:appendix}
\section{Architectures and Implementations}\label{appendix:implementation}
Our model is implemented using the PyTorch toolkit and the fairseq codebase.\footnote{\url{https://pytorch.org/};
	\url{https://github.com/pytorch/fairseq}
	}

\paragraph{Machine translation with WMT'14}
Our \base model in this experiment
is the transformer-base by~\citet{vaswani2017attention}.
Its encoder and decoder are both of 6 transformer layers.
Each multi-head attention layer is of hidden size 512, and uses 8 attention heads;
the hidden dimensions for the feed forward networks
are 2,048.
We follow issue \#346 of the fairseq's GitHub repository
to replicate the results by \citet{vaswani2017attention}.\footnote{\url{https://github.com/pytorch/fairseq/issues/346}}
When training \ourmodel, we mostly use the same hyperparameters,
with the only exception being that we warmup the learning rate for 8,000 updates,
instead of 4,000.\footnote{Due to the randomness in random expert selection,
	we find that warming up learning rate more slowly helps
	stabilize early training.}

At evaluation time, we apply early stopping based on
development set loss, and then average the most recent
5 checkpoints of the model, following~\citet{vaswani2017attention}.

\paragraph{Machine translation with IWSLT'14.}
The \base model in this experiment is due to the fairseq
codebase.\footnote{\url{https://github.com/pytorch/fairseq/tree/master/examples/translation}
	}
It mostly follows the transformer-base architecture,
but uses a larger dropout rate (0.3 vs. 0.1), a smaller feed forward network hidden size (1,024 vs. 2,048),
and a larger weight decay ($10^{-4}$ vs. 0).
We use 8,000 warmup updates.

\paragraph{Language modeling with WikiText-103.}
For the \base model, we follow the model by~\citet{baevski2018adaptive}.
The learning rate is warmed up for $240,000$ steps.

For all three experiments,
the gating functions in our \ourmodel model and the \joint baseline 
are implemented as $\tanh$-MLPs.
They have 256 hidden dimensions.
We apply a batch normalization~\citep{sergey2015batchnorm}
to the input to the MLPs.
We can see that the gating functions only have a small amount of parameters,
accounting for less than 5\% parameters of the full \ourmodel model.
A dropout of 0.1 is applied to the output of the first layer.
No weight decay is used.
$\bm{\phi}$ are updated using SGD with a fixed learning rate 1,
separate from the one for the rest part of the models.
This aims to avoid using momentum-based optimizing algorithms (e.g., Adam)
for the gating functions, which 
we empirically find 
helps alleviate the ``rich gets richer'' degeneracy.\footnote{It is not entirely clear to us why
using momentum-based optimization algorithms 
to learn the gating functions leads to degenerate solutions more often.
One possible reason is that the accumulated momentum 
steers the gating functions to
keeping selecting the experts they pick at the early stage of training.
}

In the language modeling experiment,
most recent 100 input vectors
are averaged and then fed into the gating functions;
while we average all the input vectors in the machine translation
as the inputs to $\vg(\boldsymbol{\cdot};\bm{\phi})$.

\section{Learning Curve Comparison for \ourmodel and \joint}\label{subsec:overfit}
\begin{figure}
	\centering
	\includegraphics[clip,trim=.5cm 0cm .5cm 1.0cm,width=.8\columnwidth]{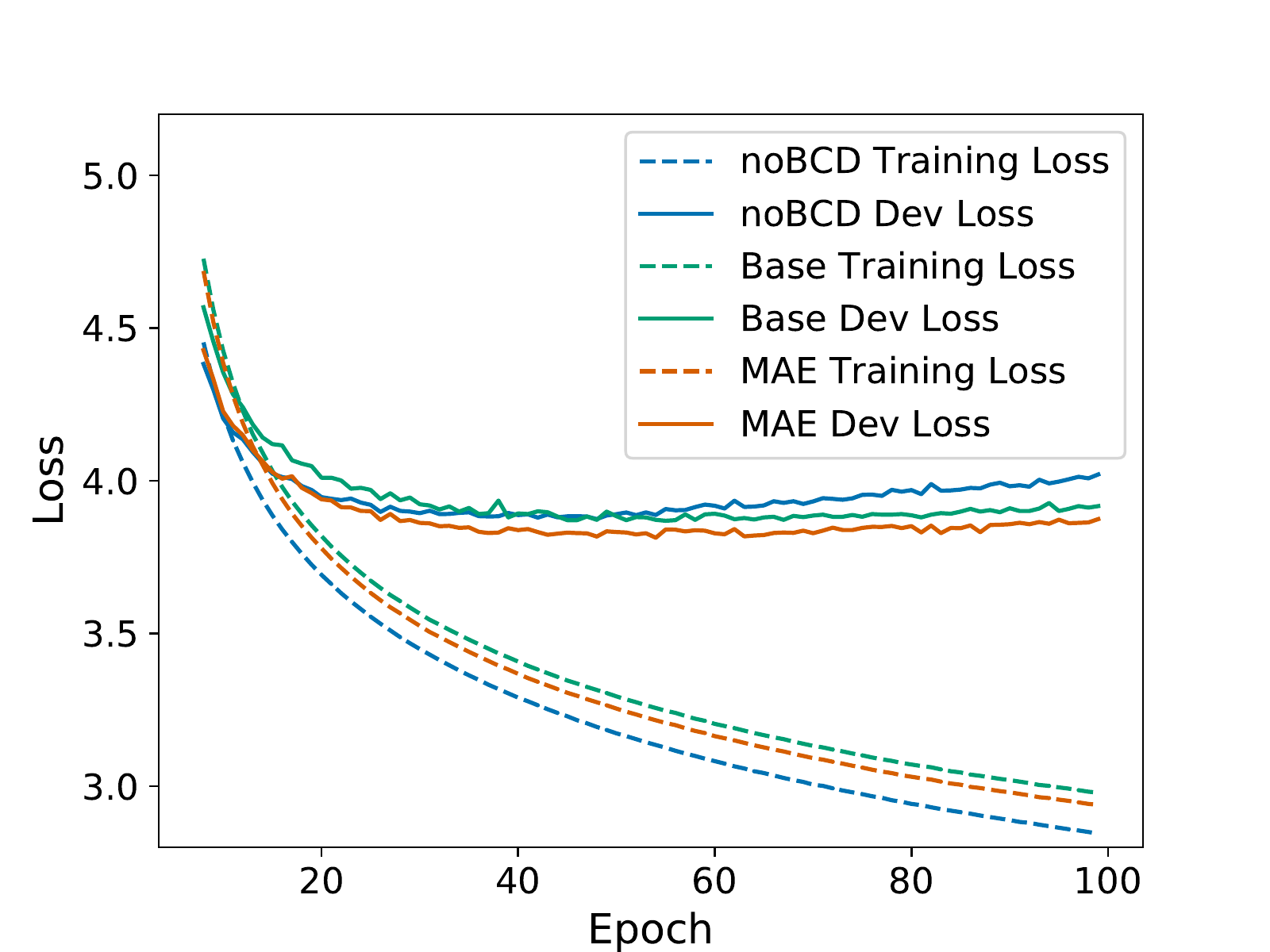}
	\caption{Learning curves of \base, \joint, and \ourmodelsc~(\S\ref{subsec:overfit}),
		trained on the IWSLT14 EN-DE using the same
		setup.
		\joint quickly fits the training data,
		but it does not outperform \base on validation set.
		Trained with BCD, \ourmodel finds a nice middle ground.
		For better readability, x-axis starts at epoch 8.
            }
	\label{fig:learning_curve}
\end{figure}

In \S\ref{sec:model} (footnote~\ref{fn:overfit}) we discuss an overfitting
issue by jointly updating the experts and the gating function.
This section empirically studies it.
We compare the learning curves of 
\base, \joint, and \ourmodelsc trained on the
IWSLT14 dataset, plotted in Figure~\ref{fig:learning_curve}.
The models are described in~\S\ref{subsec:baselines}.
We tune dropout and $\ell_2$ regularization based on development performance.
	Other hyperparameters are the same for the compared models.

The training loss for \joint decreases much faster than \base;
however, on the development set, it never 
outperforms \base,
and the development loss starts increasing after epoch 40.
\ourmodelsc finds a nice middle ground in terms of training loss.
It outperforms both \base and \joint on the validation set.
This provides further evidence for the
importance of BCD training.

\section{Addtional Results for \S\ref{subsec:cluster}}\label{subsec:attribute}
\S\ref{subsec:cluster} describes a experiment with the \ourmodelsc model
where we attribute the development instances of WMT14
to the experts maximizing the gating weights.
Table~\ref{tab:specialized_experts} presents more results.
The number of instances each expert receives is relatively balanced,
and the trend is consistent across different layers.

\begin{table}[t]
	\centering
	\small
	\begin{tabulary}{\columnwidth}{@{}c c c c c c c c c@{}}
		\toprule
		
		\textbf{Layer} & \textbf{E1} & \textbf{E2} &\textbf{E3} & \textbf{E4} & \textbf{E5} & \textbf{E6} & \textbf{E7} & \textbf{E8} \\
		\midrule[.03em]
		1 & 13.1 & 13.9 & 8.9 & 16.1 & 10.3 & 15.3 & 10.1 & 11.6\\
		2 & 13.8 & 14.5 & 10.7 & 10.8& 15.4 & 7.9 & 16.0 & 10.9\\
		3 & 14.0 & 14.4 & 12.4 & 10.6 & 14.3 & 9.8 & 15.4 & 9.0\\
		4 & 14.5 & 13.7  & 10.4 & 8.3 & 15.1 & 11.8 & 11.2 & 15.1\\
		5 & 11.9 & 13.8 & 13.7 & 15.7 & 10.1 & 16.4 & 6.9 & 11.5\\
		6 & 12.9 & 10.0 & 12.4 & 14.6 & 9.5 & 15.2 & 15.7 & 9.8\\
		
		\bottomrule
	\end{tabulary}
	\caption{The percentage of WMT14 development instances attributed to each of the experts in \ourmodelsc's encoder layers~(\S\ref{subsec:cluster}).}
	\label{tab:specialized_experts}
\end{table}
\end{appendices}

\end{document}